\documentclass{article}

\usepackage{microtype}
\usepackage{graphicx}
\usepackage{subfigure}
\usepackage{booktabs} % for professional tables
\usepackage[hyphens]{url}
\usepackage{breakurl}
\usepackage{enumitem}
\usepackage{scalerel}

\usepackage{hyperref}

\usepackage{dashbox}
\usepackage[
    n,
    operators,
    advantage,
    sets,
    adversary,
    landau,
    probability,
    notions,    
    ff,
    mm,
    primitives,
    events,
    complexity,
    asymptotics,
    keys]{cryptocode}

\usepackage[accepted]{icml2025}

\usepackage{amsmath}
\usepackage{amssymb}
\usepackage{mathtools}
\usepackage{amsthm}
\usepackage{tcolorbox}

\usepackage[capitalize,noabbrev]{cleveref}

\theoremstyle{plain}

\theoremstyle{definition}

\theoremstyle{remark}

\usepackage{enumitem}
\usepackage{diagbox}
\usepackage{makecell}
\usepackage{float}
\usepackage{threeparttable}
\usepackage{placeins}
\usepackage{listings}
\usepackage{ebproof}

\usepackage[textsize=tiny]{todonotes}

\definecolor{red}{RGB}{220,40,30}
\definecolor{BrightOrange}{RGB}{255,140,0}
\definecolor{Gold}{RGB}{255,200,0}
\definecolor{Emerald}{RGB}{0,160,80}
\definecolor{Cobalt}{RGB}{0,120,215}
\definecolor{Sapphire}{RGB}{90,60,170}
\definecolor{Amethyst}{RGB}{160,80,200}
\definecolor{framegray}{RGB}{170,170,170}
\newcommand{\bench}{\textsc{IntegralBench}}

\begin{document}

\twocolumn[
\icmltitle{\bench: Benchmarking LLMs with Definite Integral Problems}



\icmlsetsymbol{equal}{*}

\begin{icmlauthorlist}
\icmlauthor{Bintao Tang}{equal,tongji}
\icmlauthor{Xin Yang}{equal,zju}
\icmlauthor{Yuhao Wang}{equal,tongji}
\icmlauthor{Zixuan Qiu}{xju}
\icmlauthor{Zimo Ji}{hkust}
\icmlauthor{Wenyuan Jiang}{eth}
\end{icmlauthorlist}

\icmlaffiliation{tongji}{School of Software Engineering, Tongji University, Shanghai, China}
\icmlaffiliation{zju}{Polytechnic Institute, Zhejiang University, Zhejiang, China}
\icmlaffiliation{eth}{ETH Zurich, Zurich, Switzerland}
\icmlaffiliation{hkust}{Department of Computer Science and Engineering, Hong Kong University of Science and Technology, Clear Water Bay, Hong Kong}
\icmlaffiliation{xju}{School of Mathematics and Physics, Xi’an Jiaotong Liverpool University, Suzhou, China }

\icmlcorrespondingauthor{Wenyuan Jiang}{wenyjiang@ethz.ch}
\icmlcorrespondingauthor{Zimo Ji}{zjiag@connect.ust.hk}

\icmlkeywords{Machine Learning, ICML}

\vskip 0.3in
]



\printAffiliationsAndNotice{\icmlEqualContribution} % otherwise use the standard text.
\setlist[itemize]{noitemsep, topsep=0pt}

\begin{abstract}
We present \bench, a focused benchmark designed to evaluate Large Language Model (LLM) performance on definite integral problems.
\bench~provides both symbolic and numerical ground truth solutions with manual difficulty annotations. 
Our evaluation of nine state-of-the-art LLMs reveals significant performance gaps and strong correlations between problem difficulty and model accuracy, establishing baseline metrics for this challenging domain. 
\bench~aims to advance automated mathematical reasoning by providing a rigorous evaluation framework specifically tailored for definite integral computation.
\end{abstract}

\section{Introduction}

Mathematical reasoning represents one of the most advanced forms of human intelligence and serves as a critical benchmark for evaluating Large Language Model (LLM) capabilities. 
Several benchmarks currently assess LLMs' mathematical performance:
MATH~\cite{hendrycksmath2021} tests advanced high school competition problems, GSM8K~\cite{cobbe2021training} focuses on grade school arithmetic word problems, and MathVista~\cite{lumathvista} evaluates multimodal mathematical reasoning. 
This widespread attention underscores the recognized importance of mathematical evaluation for LLMs.

Within this domain, definite integral problems offer a uniquely challenging testbed for assessing both computational accuracy and symbolic reasoning. 
Unlike elementary arithmetic or algebra, integral calculus demands sophisticated multi-step reasoning including decomposition of complex expressions, pattern recognition for simplification techniques, and recall of integration methods. 
These characteristics make integral calculus particularly suitable for evaluating advanced LLM reasoning capabilities.

Despite existing mathematical benchmarks, current evaluation frameworks exhibit significant limitations for integral problems.
First, most benchmarks contain insufficient integral problems for meaningful assessment. 
While MATH includes calculus problems, it has relatively few challenging integrals that comprehensively test integration techniques. 
Second, current frameworks lack metrics specifically designed for integral evaluation, such as differentiating between symbolic and numerical solution accuracy. 
Finally, existing benchmarks rarely implement appropriate difficulty gradation for integrals, failing to distinguish between routine applications and problems requiring advanced techniques. 
This lack of stratification restricts the precise measurement of model capabilities across complexity levels.

\begin{figure}[t]
    \centering
    \begin{pcvstack}[boxed, space=0.2cm]
    \procedureblock{}{%
    \textbf{\normalsize Problem} \\ [][\hline]
    \\ [-0.3cm]
\scaleto{\int\limits_0^1\frac1{\sqrt{1+4x^2}}\ln\left(2x+\sqrt{1+4x^2}\right)\arccos x\:dx}{38pt}\\
    \\ [-0.3cm]
    \textbf{\normalsize Answer}\\ [][\hline]
    \\ [-0.3cm]
    \textsf{Symbolic} = -\frac{\pi}{16}\mathrm{Li}_2(-4)\\
    \textsf{Numerical} = 0.465336591\\
    \\ [-0.3cm]
    \textbf{\normalsize Difficulty}\\ [][\hline]
    \\ [-0.3cm]
    \textsf{Rating} = 4 \bigstar\\
    \\ [-0.3cm]
    \textbf{\normalsize Source}\\ [][\hline]
    \\ [-0.3cm]
    \text{Brychkov, Yury A. \textit{Handbook of Special Functions.} CRC Press,}\\
    \text{28 May 2008, p. 166.}
    \\ [-0.7cm]
    } 
    \end{pcvstack}
    % \vspace{-0.7cm}
    % \caption{Example of a challenge in \bench consisting of problem, symbolic/numerical answer, difficulty, and problem source.}
    \caption{Example problem from \bench~with symbolic/numerical ground truth solutions, difficulty rating, and source attribution.}
    \label{fig:example-data}
    % \vspace{-0.5cm}
\end{figure}

To address these limitations, we introduce \bench, a focused benchmark specifically designed for evaluating LLM performance on definite integral problems. 
\bench~comprises 317 carefully selected graduate-level problems sourced from advanced textbooks and competitions. 
Each problem provides both symbolic and numerical ground truth solutions as is shown in Figure \ref{fig:example-data}, enabling separate assessment of LLM-generated answers through distinct evaluation metrics.

Additionally, each problem is manually annotated with difficulty ratings from 1 to 5, enabling fine-grained analysis across varying complexity levels. 
\bench~also incorporates a novel term-rewriting method to generate problem variations, preventing dataset contamination while maintaining mathematical rigor.
In terms of construction cost, \bench~employs a systematic methodology that balances the trilemma of cost, difficulty, and relevance for building benchmark datasets through LLM-assisted curation from academic sources to create challenging mathematical benchmarks.

Our evaluation of nine state-of-the-art LLMs yields several key insights. Larger models generally perform better, with Qwen3-235B-A22B~\cite{qwen3} achieving the highest accuracy—50.16\% on numerical solutions and 56.15\% on symbolic solutions. 
However, model size alone does not determine performance; the 32B QwQ~\cite{qwq32b} model outperforms larger models including GPT-4.1~\cite{GPT-4.1} and Claude 3.7~\cite{anthropic2024claude37}, demonstrating the importance of architecture and training methodology.

We observe a strong negative correlation between problem difficulty and model accuracy across all evaluated models, with performance declining sharply on challenging problems. 
This validates our difficulty annotations and reveals current limitations in complex mathematical reasoning.

Our analysis of inference-time scaling shows that accuracy improves rapidly during initial token consumption before plateauing, with different models exhibiting distinct ``sweet spots.'' 
This suggests varying efficiency in information extraction during extended reasoning.

In summary, our contributions are threefold:
\begin{itemize}
\item \textbf{Dataset:} We construct \bench, a focused benchmark of 317 graduate-level integral problems with verified solutions for evaluating advanced LLM mathematical reasoning. 
\footnote{The dataset is publicly available at \url{https://github.com/vegetable-yx/IntegralBench/}.}
% The dataset and evaluation scripts are publicly available at \textcolor{blue}{\href{https://anonymous.4open.science/r/IntegralBench-32C7/}{Anonymous GitHub}}.
\item \textbf{Pipeline:} We propose a scalable methodology for constructing challenging mathematical benchmarks through LLM-assisted curation from academic sources, providing a framework for future benchmark development.
\item \textbf{Evaluation:} We conduct a comprehensive evaluation of nine mainstream LLMs on \bench, revealing strengths and limitations in definite integral computation and informing future research directions.
\end{itemize}
\section{The \bench~Dataset}

\begin{figure}[htb]
    \centering
    \includegraphics[trim={0 0.2cm 0 0},clip,width=0.5\textwidth]{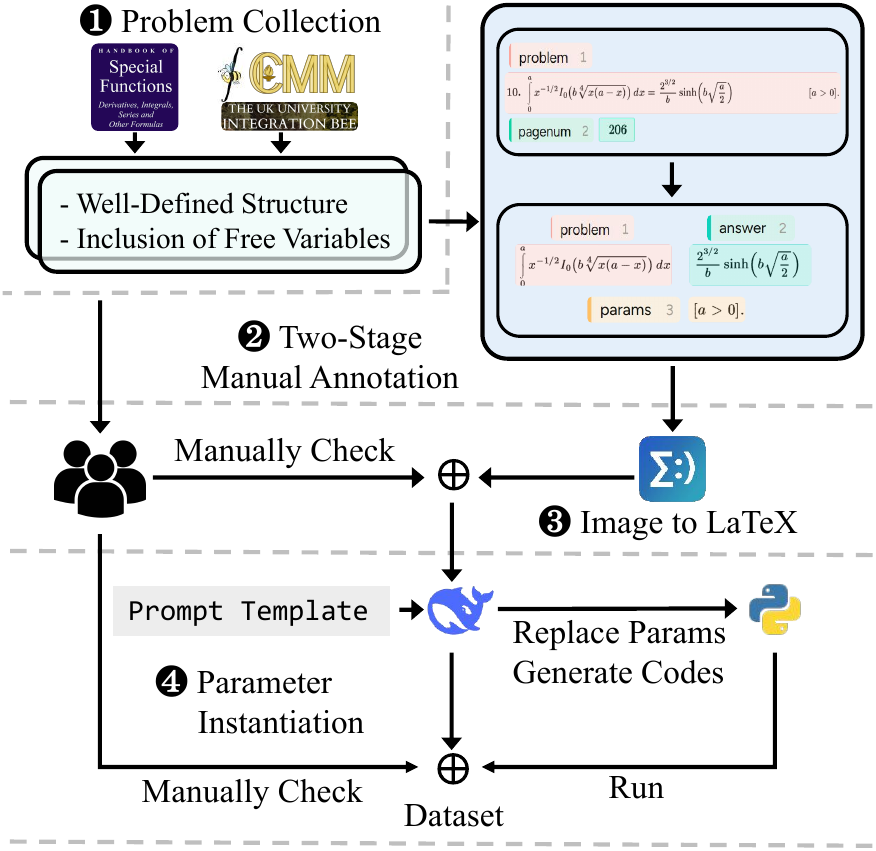}
    % \vspace{-0.7cm}
    \caption{Construction pipeline for \bench. The workflow consists of four steps: (1) problem collection from graduate textbooks and competitions with selection criteria, (2) two-stage manual annotation using bounding boxes, (3) OCR-based conversion to LaTeX with manual verification, and (4) parameter instantiation for problems with free variables, followed by final dataset validation.}
    \label{fig:overview}
\end{figure}

We start with the definition of definite integrals in the context of our dataset. 
Then, we explain how \bench~is created and present the features of this dataset. 
% The dataset will be publicly available upon the acceptance of the paper.

\subsection{Definite Integrals}
Informally, definite integrals represent the area under a curve between specified bounds, capturing fundamental concepts in analysis and applied mathematics. Formally, a definite integral is defined as:
$$
\int_a^b f(x) \, dx = \lim_{n \to \infty} \sum_{i=1}^n f(x_i^*) \Delta x
$$
where $a$ and $b$ are the integration limits, $f(x)$ is the integrand, and the limit represents the Riemann sum as partition width approaches zero. 
Examples of definite integral expressions are shown in Figure~\ref{fig:example-data}. We refer to these complete expressions as \emph{integral bodies} throughout this work. Each definite integral body consists of three essential components: (1) the \emph{integrand} (the function being integrated), (2) the \emph{integration limits} (bounds of integration), and (3) the \emph{integration variable}. Notably, an integrand may contain composite functions, special functions, or even nested integrals, leading to substantial complexity. Furthermore, integral bodies may contain parameters or \emph{free variables} that are not bound by the integration operation.

\begin{figure*}[htbp]
    \centering
    \includegraphics[width=1.0\textwidth]{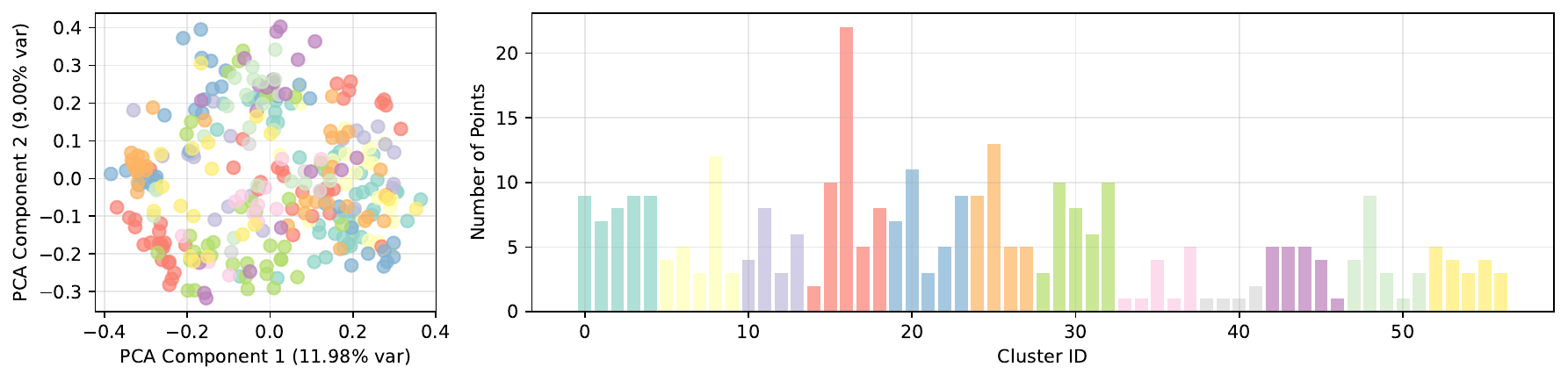}
    % \vspace{-0.3cm}
        \caption{Diversity analysis of \bench~ using affinity propagation clustering on problem-solution embeddings. Left: PCA visualization of 56 distinct clusters showing well-distributed coverage across the embedding space. Right: Cluster size distribution reveals a balanced representation with most clusters containing 3-13 problems, avoiding over-concentration in any single problem type.}
    \label{fig:clusters_diversity}
\end{figure*}

\subsection{Creating \bench~}

The workflow to create \bench~ is shown in Figure~\ref{fig:overview}.
We first select problem sources containing challenging definite integral problems, then sample problems according to inclusion criteria.
Sampled problems are manually annotated with ground truth answers and metadata, and converted to LaTeX using OCR, with parameter instantiation for problems containing free variables.
The resulting dataset undergoes manual inspection to ensure correctness.

\noindent\textbf{Problem sources.} 
We collect definite integral problems from two publicly available sources: (1) graduate-level textbooks and (2) integral competitions.
We focus on integral computations involving complex expressions with elementary and special functions, excluding theoretical proofs typical of analysis textbooks.
After manual inspection, we retain only graduate and advanced undergraduate-level sources that provide ground truth solutions.
This process yields one textbook and three integral competition series as our final data sources.

\noindent\textbf{Selection criteria.} 
We sample problems using exclusion rules where any problem meeting a rule is excluded from the dataset.
The criteria enforce: (1) problems must evaluate valid definite integrals with all components present; (2) subexpressions must be well-formed with reasonable structure involving elementary and special functions; (3) ground truth must be numerically computable in constant time.

\noindent\textbf{Manual annotation.} 
We employ a two-stage annotation pipeline using Label Studio.
To preserve mathematical typesetting accuracy, we curate problem images from PDF files rather than direct text extraction.
In stage one, we annotate bounding boxes for problems satisfying selection criteria, creating data points $(x_0,y_0,x_1,y_1,\textit{source\_image})$ where coordinates define the bounding box and \textit{source\_image} is the source PDF page.
In stage two, we annotate bounding boxes for integral components (body, answer, parameter constraints), resulting in $((x_0^i,y_0^i,x_1^i,y_1^i)_{i\in\{\text{body},\text{ans},\text{param}\}},\textit{source\_image})$.
We exclude problems with multi-line expressions or overlapping bounding boxes to maintain annotation quality.

\noindent\textbf{Image to LaTeX conversion.} 
We convert annotated images to LaTeX using SimpleTex V2.5~\cite{simpletexSimpleTexSnip}.
Each bounding box-image pair produces OCR input, with outputs validated through regular expression checks for well-formed LaTeX.
All LaTeX-image pairs undergo manual verification to ensure correctness, with incorrect formulas marked and manually corrected.

\noindent\textbf{Parameter instantiation.} 
Problems with free variables lack closed-form solutions and exact numerical values.
We instantiate free variables with concrete values based on their constraints.
DeepSeek V3 generates Python scripts that randomly sample parameter values and substitute them into integral expressions and answers.
Each original-instantiated problem pair undergoes manual verification to ensure correctness.

\noindent\textbf{Ground truth solutions.} 
Ground truth comprises symbolic and numerical components.
Symbolic solutions derive from parameter-instantiated expressions after LLM-assisted simplification.
Numerical solutions result from executing parameter-instantiated computational scripts.
Both solution types undergo manual quality control inspection.

\noindent\textbf{Difficulty.}
Human experts were invited to carefully assess each problem in our dataset. Each problem was assigned a difficulty rating on a scale from 1 (easiest) to 5 (most difficult), with higher scores indicating increased difficulty.

\subsection{Diversity and Cost}
The resulting dataset contains 317 integral problems with ground truth answers.
We analyze the diversity of mathematical concepts of the resulting dataset and the cost of building the dataset.

\noindent\textbf{Diversity.} 
To measure the diversity of integral problems, we performed affinity propagation clustering~\cite{frey2007clustering} over embeddings of each problem-solution pair.
We used Qwen3-Embedding-0.6B~\cite{qwen3embedding} as the embedding model.
The damping factor for affinity propagation was set to 0.5 with maximum iterations of 200, terminating early if no changes occurred across 15 consecutive iterations.
This resulted in 56 distinct clusters with sizes shown in Figure~\ref{fig:clusters_diversity}.

The clustering analysis reveals substantial diversity in \bench~problems. 
The PCA visualization shows well-distributed clusters across the embedding space, indicating that problems cover diverse mathematical concepts rather than concentrating on a few similar areas. 
The cluster size distribution demonstrates a healthy spread: while the largest cluster contains 22 problems, most clusters are smaller (3-13 problems), suggesting that \bench~avoids over-representation of any single problem type. 
The presence of numerous small clusters indicates coverage of specialized integration techniques and mathematical structures, supporting \bench's suitability for comprehensive evaluation of integral reasoning capabilities.

\noindent\textbf{Cost.} 
Since we adopt an LLM-heavy method for dataset curation with human quality control, the two major parts of the cost are for the tokens of LLM and human working hours in checking.
For LLM-related costs, we recorded the token count and per token cost of each LLM call based on the statistics from API providers.
Human working hours are collected from Label Studio by adding up time for each annotation task.
The cost is estimated using the \$32 per hour rate for the average salary of a Graduate Research Assistant in the US.
We summarize each category and the total cost in Table \ref{tab:dataset-cost}.

\begin{table}[htb]
    \centering
    \begin{tabular}{llr}
        \toprule
        \textbf{Category} & & \textbf{Cost}\\
        \midrule
        API    & DeepSeek             &  0.83\$\\
               & SimpleTex            &  4.17\$\\
        \midrule
        Human & Stage 1 + Check       & 21.03\$\\
              & Stage 2 + Check       & 210.86\$\\
              & Difficulty annotation & 248.04\$\\
        \midrule
        \textbf{Total} & & 484.93\$\\
        \bottomrule
    \end{tabular}
    \caption{
    Cost breakdown for constructing \bench~dataset. Human annotation comprises the majority of costs (\$479.93), while API calls contribute minimally (\$5.00).
    }
    \label{tab:dataset-cost}
\end{table}

\section{Experiments}

\begin{figure}[t]
    \centering
    \includegraphics[trim={0 0.2cm 0 0.2cm},clip,width=0.45\textwidth]{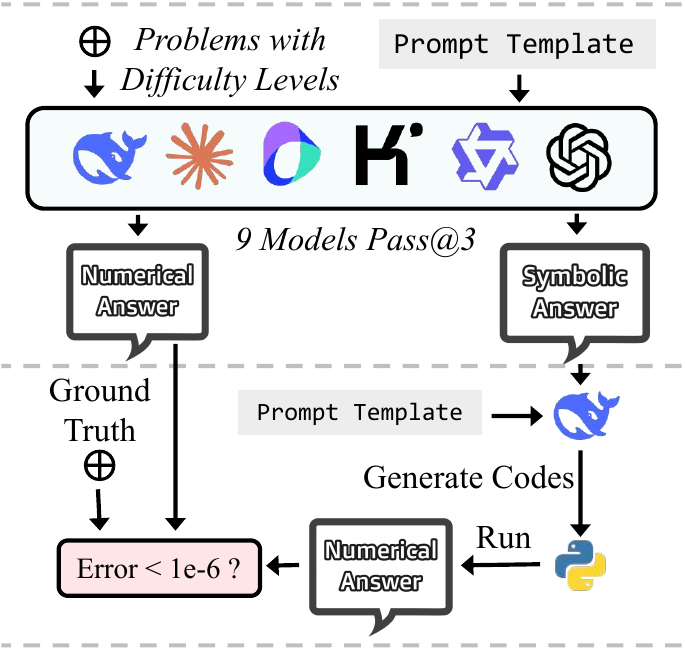}
    % \caption{Pipeline of evaluating on \bench.}
    \caption{Evaluation pipeline for \bench. Nine LLMs generate both numerical and symbolic answers using standardized prompts. Numerical answers are validated by direct comparison with ground truth (error $< 10^{-6}$), while symbolic answers are verified through LLM-generated code execution to ensure consistency with the provided numerical result.}
    \label{fig:eval_pipeline}
\end{figure}

We evaluate the performance of large language models on our \bench~ in this section and present the corresponding results and findings. Figure~\ref{fig:eval_pipeline} illustrates our evaluation pipeline.

\begin{table*}[htb]
\centering
\begin{tabular}{lccccc} 
\toprule
\textbf{Model} & \textbf{Model Size} & \textbf{PASS@3(Num)} & \textbf{ALL@3(Num)} & \textbf{PASS@3(Sym)} & \textbf{ALL@3(Sym)} \\
\midrule
\textbf{DeepSeek-V3}   & 671B & 27.44\% & 17.98\% & 35.96\% & 22.71\% \\
\textbf{DeepSeek-R1}   & 671B & 45.43\% & 29.97\% & 53.63\% & 37.85\% \\
\textbf{Doubao 1.5 thinking pro} & 200B & 45.43\% & 30.60\% & 52.37\% & 40.38\% \\
\textbf{Qwen3-235B-A22B}   & 235B & 50.16\% & 28.71\% & 56.15\% & 38.17\% \\
\textbf{QwQ-32B}   & 32B & 44.48\% & 30.28\% & 49.21\% & 36.91\% \\
\midrule
\textbf{GPT-4.1}   & / & 26.81\% & 18.93\% & 47.00\% & 26.81\% \\
\textbf{O3-mini} & / & 35.65\% & 25.55\% & 43.35\% & 33.86\% \\
\textbf{Kimi-K1.5}     & / & 22.40\% & 13.88\% & 28.71\% & 17.35\% \\
\textbf{Claude 3.7} & / & 24.61\% & 14.51\% & 29.97\% & 16.72\%  \\
\bottomrule
\end{tabular}
\caption{Model performance on \bench~using PASS@3 and ALL@3 metrics. Num (Numerical) evaluates predicted numerical answers; Sym (Symbolic) evaluates analytical expressions. PASS@3 requires correct solutions in at least one of three attempts; ALL@3 requires correct solutions in all three attempts. Parameter sizes are undisclosed for some models.}
\label{tab:model_type_size_comparison}
\end{table*}

\subsection{Evaluation settings}

\noindent\textbf{Evaluated models.}
We evaluate a total of 9 large language models, covering a range of open-source and proprietary systems: Claude 3.7~\cite{anthropic2024claude37}, Doubao 1.5 thinking pro~\cite{Seed-Thinking-v1.5}, GPT-4.1~\cite{GPT-4.1}, O3-mini~\cite{o3-mini}, DeepSeek-V3~\cite{deepseekai2024deepseekv3technicalreport}, DeepSeek-R1~\cite{deepseekai2025deepseekr1incentivizingreasoningcapability}, Kimi-K1.5~\cite{team2025kimi}, Qwen3-235B-A22B~\cite{qwen3}, QwQ-32B\cite{qwq32b}. These models are selected to ensure diversity in architecture, training corpus, and reasoning capabilities. 
The chosen models are representative of current state-of-the-art LLMs in mathematical reasoning and symbolic computation.

\noindent\textbf{Inference setting.}
Given the challenging nature of problems requiring multi-step symbolic reasoning, we set the maximum output length (\textit{max\_tokens}) to each model's maximum capacity to prevent premature truncation of solutions. To ensure fair comparison across models, we use default temperature and top-p values during inference, preserving each model's intended decoding behavior. 

\noindent\textbf{Metrics.} To account for generation variability and evaluate robustness, each problem is tested $N=3$ times independently per model. This enables the computation of two key metrics: \textit{PASS@N}, which measures whether the model produces the correct solution in at least one of the $N$ attempts, and \textit{ALL@N}, which requires correct solutions in all $N$ attempts. These complementary metrics assess both the model's peak performance capability and its consistency in mathematical reasoning.

\noindent\textbf{Answer validation.}
The correctness of model outputs is assessed by verifying both the symbolic and numerical components:
\begin{itemize}
    \item \textbf{Numerical answer verification:} We compare the model-provided \textit{numerical\_answer} to the ground truth. If the absolute error is below $10^{-6}$, the result is marked correct.
    \item \textbf{Symbolic answer verification:} The model-generated answer is passed to DeepSeek, which generates Python code to numerically evaluate it. If this computed result matches the model's numerical\_answer within the $10^{-6}$ threshold, the symbolic answer is considered valid.
\end{itemize}
This dual-checking mechanism ensures that both symbolic reasoning and numerical consistency are taken into account in the evaluation.

\noindent\textbf{Prompts.}
To ensure a fair and consistent evaluation, all models are prompted using a standardized template for definite integral problems. 
The prompt template can be found in the Appendix. 
This prompt ensures that models produce both a symbolic solution and a numerical approximation, following a structured reasoning chain.

\begin{figure*}[htb]
    \centering
    \includegraphics[trim={0 0 0 0},clip,width=1\textwidth]{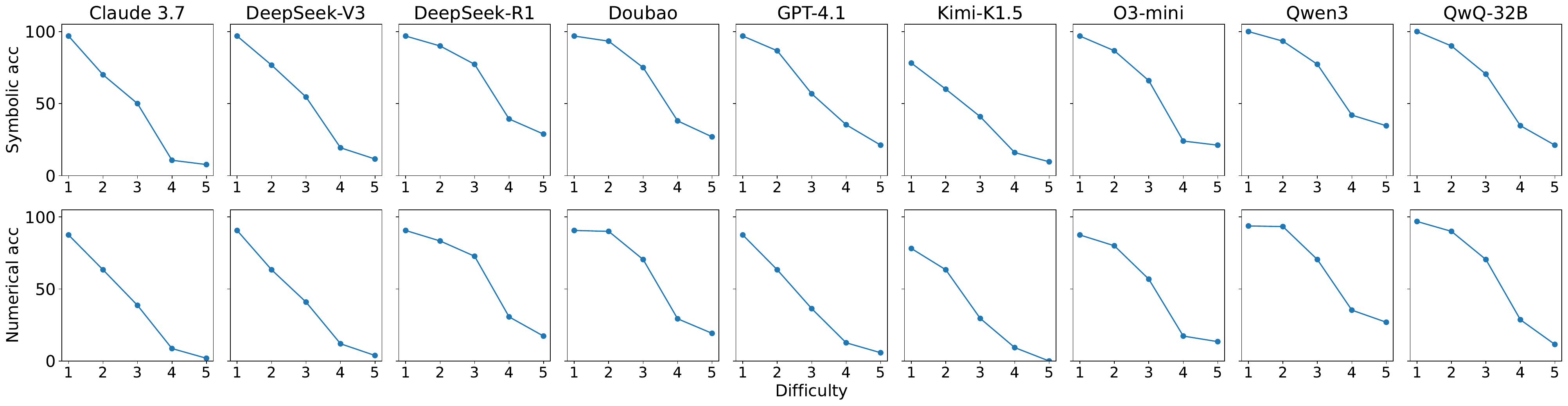}
    % \vspace{-0.5cm}
    \caption{PASS@3 performance across difficulty levels for symbolic and numerical evaluations. All models show consistent performance decline with increasing difficulty, demonstrating a strong negative correlation between problem complexity and accuracy.}
    \label{fig:difficulty_result}
\end{figure*}

\subsection{RQ1: Model type vs. Performance}

We examine how different model architectures and sizes perform on numerical and symbolic tasks. As shown in Table~\ref{tab:model_type_size_comparison}, larger models generally achieve better results, with Qwen3-235B-A22B leading overall at 50.16\% PASS@3 numerical and 56.15\% PASS@3 symbolic accuracy. 
DeepSeek-R1 and Doubao 1.5 Thinking Pro also demonstrate strong performance across both evaluation types.

Notably, the 32B QwQ model performs surprisingly well, outperforming larger models including GPT-4.1 and Claude 3.7. 
This suggests that architecture design and training methodology significantly impact performance beyond raw parameter count. 
Models with undisclosed sizes show varied performance, with O3-mini achieving moderate results while Kimi-K1.5 and Claude 3.7 show lower accuracy across all metrics.

\begin{tcolorbox}
\textbf{Finding 1:} 
While model size generally correlates with performance, architecture, and training methodology are equally critical factors. 
The 32B QwQ model outperforms several larger models, demonstrating that parameter count alone does not determine mathematical reasoning capability.
\end{tcolorbox}

\subsection{RQ2: Problem difficulty vs. Performance}

We analyze the relationship between manually annotated problem difficulty (rated 1-5) and model performance. 
Figure~\ref{fig:difficulty_result} reveals a strong negative correlation between difficulty and PASS@3 accuracy across all evaluated models. 
While models achieve near-perfect performance on easier problems (difficulty 1-2), accuracy drops dramatically on the most challenging problems (difficulty 4-5), often approaching zero.

This consistent pattern validates our difficulty annotations and demonstrates that \bench~effectively spans a comprehensive range of complexity levels. 
The sharp performance decline on difficult problems reveals current limitations in LLM mathematical reasoning capabilities, particularly for problems requiring advanced integration techniques or multi-step symbolic manipulation.

\begin{tcolorbox}
\textbf{Finding 2:} 
All models exhibit a strong negative correlation between problem difficulty and accuracy, with performance dropping sharply on challenging problems. 
This validates our difficulty stratification and reveals current limitations in complex mathematical reasoning across state-of-the-art models.
\end{tcolorbox}

\subsection{RQ3: Inference-time scaling vs. Performance}

\begin{figure}[htb]
    \centering
    \includegraphics[width=0.45\textwidth]{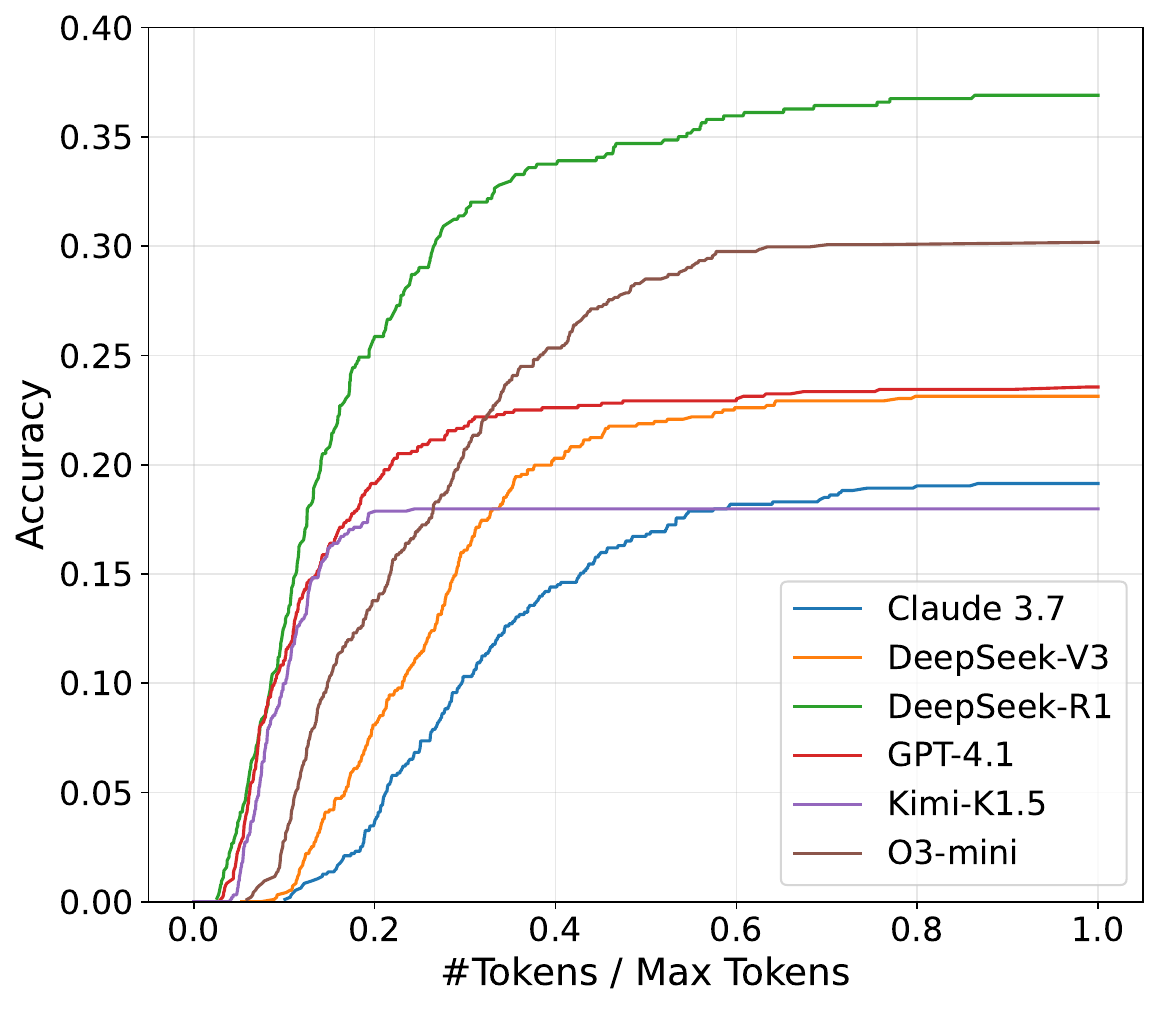}
    % \vspace{-0.3cm}
    \caption{
    Inference-time scaling curves showing normalized token consumption versus cumulative PASS@3 accuracy. 
    Most models exhibit rapid early gains followed by saturation, with varying ``sweet spot" locations highlighting differences in inference efficiency across models.
    }
    \label{fig:token_acc}
\end{figure}

We investigate how inference-time token consumption affects model performance by analyzing accuracy curves across token usage ratios. 
As shown in Figure~\ref{fig:token_acc}, all models exhibit rapid accuracy gains during initial token consumption, followed by plateauing behavior after reaching model-specific ``sweet spots".

The location of these plateaus varies significantly across models. Kimi-K1.5 reaches peak performance around 20\% token ratio, while DeepSeek-R1 continues showing marginal improvements beyond 80\% token usage. 
This suggests fundamental differences in reasoning efficiency: some models extract crucial information early in generation, while others benefit from extended reasoning chains.

\begin{tcolorbox}
\textbf{Finding 3:} 
Models show rapid early accuracy gains followed by plateauing, with model-specific ``sweet spots" varying significantly. 
This reveals different inference efficiency patterns—some models might have been optimizing for early information extraction while others can benefit from extended reasoning token budgets.
\end{tcolorbox}

\subsection{Failure mode analysis}
We conducted a detailed analysis of incorrect cases and identified several representative categories of typical failure modes. Full examples of each error type are provided in the appendix to facilitate a clearer understanding of their characteristics.

\noindent\textbf{Output truncation.} 
Models sometimes generate excessively verbose reasoning steps, causing truncation before reaching final answers. 
This occurs when models attempt overly complex solution paths rather than identifying efficient approaches. 
The verbose generation suggests poor strategic planning in mathematical problem-solving.

\begin{lstlisting}[breaklines]
### Step 2: Numerical Approximation
...
Higher terms decay rapidly. Summing the first few terms and multiplying by 
[Stopped here.]
\end{lstlisting}
For example, in the case above, the generation reaches the token limit during the numerical approximation phase, failing to provide the final answer despite having established the correct analytical framework.

\noindent\textbf{Circular reasoning patterns.} 
Models occasionally enter infinite loops, repeating identical expressions or computations. 
This represents a more fundamental generation failure than simple truncation, suggesting deficiencies in the model's internal state management during complex mathematical derivations.

\begin{lstlisting}[breaklines]
Simplifying the powers of 2:
...
= 2^{(k+1)/2 + k/2 + 1} = 2^{(k+1)/2 + k/2 + 1} = 2^{(k+1)/2 + k/2 + 1}
[Repeating ' = 2^{(k+1)/2 + k/2 + 1}'.]
\end{lstlisting}

In the example above, during the algebraic simplification of exponential expressions, the model correctly identifies the need to combine exponents but becomes trapped in a repetitive pattern where it repeatedly writes the same expression without progressing toward the simplified form $2^{k+2}$. 
This indicates a failure in the model's ability to recognize when a computational step has been completed and to advance to the next stage of the derivation.

\noindent\textbf{Format violations.} 
Models produce mathematically plausible but incorrectly formatted responses that fail automated parsing. 
This indicates limitations in adhering to structured output requirements while maintaining mathematical correctness.

\begin{lstlisting}[breaklines]
### Step 3: Final Answer in JSON Format

\[
\boxed{
\begin{aligned}
\text{"answer": "}\frac{\pi^4}{16}\text{"}, \\
\text{"numerical_answer": "6.0880681896"}
\end{aligned}
}
\]
\end{lstlisting}

The model correctly computes the symbolic answer $\frac{\pi^4}{16}$ and its numerical approximation but embeds the JSON response within LaTeX mathematical environments and formatting commands in the example above. 
While the mathematical content is possibly correct, the output fails to conform to the required JSON schema, preventing automated evaluation despite containing the correct solution.

\noindent\textbf{Refusal to provide symbolic answers.} 
Models occasionally refuse to provide symbolic solutions, stating ``No simple closed-form solution" even for problems with known symbolic answers. 
This conservative behavior reveals uncertainty in handling complex symbolic expressions and may indicate training biases toward avoiding difficult mathematical tasks.

\begin{lstlisting}[breaklines]
```json
{
  "answer": "\\text{No simple closed-form solution}", 
  "numerical_answer": "1.7724538509"
}
```
\end{lstlisting}

In the example above, the model provides an accurate numerical result but claims no symbolic solution exists. 
This overly conservative approach suggests the model prioritizes avoiding incorrect symbolic expressions over attempting to derive valid analytical results, potentially reflecting training emphasis on numerical methods rather than symbolic computation.

\noindent\textbf{Symbolic-numerical inconsistency.} 
The most prevalent failure mode involves correct symbolic solutions paired with incorrect numerical evaluations. 
Table~\ref{tab:model_type_size_comparison} consistently shows higher symbolic than numerical accuracy across all models, indicating systematic weaknesses in numerical computation despite strong symbolic reasoning capabilities. 
This suggests that while models excel at symbolic manipulation, they struggle with accurate numerical evaluation of complex expressions, highlighting the need for external computational verification tools.

\section{Limitations \& Risks}

\noindent\textbf{Human verification.} 
While \bench~leverages LLM-based methods for construction compared to manually curated datasets, the benchmark still requires human expert verification to ensure correctness and quality. 
This introduces scalability and reliability limitations, as manual verification is time-consuming and inherently error-prone compared to fully automated approaches using formal verification and proof assistants. 
The human verification bottleneck may limit our ability to rapidly expand the benchmark or guarantee complete accuracy across all problems.

\noindent\textbf{LLM inference randomness.}
Our experiments demonstrate that for challenging benchmarks like \bench, the performance gap between PASS@3 and ALL@3 is substantial. 
This indicates that stochastic sampling during LLM inference introduces significant variability in benchmark results, potentially masking true model capabilities. 
Higher sampling rates (e.g., PASS@16) would provide more robust estimates for both PASS@N and ALL@N metrics, better characterizing the range of LLM performance under inference randomness.

\noindent\textbf{Numerical stability.} 
The evaluation of LLM responses relies on numerical computations involving floating-point arithmetic for both direct answer verification and intermediate calculations. 
Floating-point numerical stability has long been recognized as problematic, particularly for operations that can amplify rounding errors. 
In the context of integral evaluation, certain problem classes may exhibit poor numerical conditioning where even adaptive numerical integration algorithms cannot guarantee stable results, potentially leading to false negatives in our evaluation pipeline.

\section{Related Work}

Our work intersects several research directions in mathematical reasoning, benchmarking, and LLM evaluation. We review the most relevant literature below.

\noindent\textbf{AI for mathematics.} Recent advances have significantly improved LLM mathematical capabilities through models like Llemma~\cite{azerbayev2023llemma}, WizardMath~\cite{luo2023wizardmath}, and MAmmoTH~\cite{yue2023mammothbuildingmathgeneralist}, which achieve strong performance on benchmarks such as MATH~\cite{hendrycks2021measuring} and GSM8K~\cite{cobbe2021trainingverifierssolvemath}. Multimodal datasets like MathVista~\cite{lu2024mathvista} and GeomVerse~\cite{kazemi2023geomversesystematicevaluationlarge} further evaluate visual mathematical reasoning. However, these efforts primarily target general mathematical skills with a limited focus on specialized domains like integral calculus, lacking the depth needed for rigorous evaluation of advanced reasoning in particular subfields.

\noindent\textbf{Mathematical benchmarking.} Several benchmarks incorporate integral problems, including MathBench~\cite{liu2024mathbench} and MathVista~\cite{lu2024mathvista}, but typically peak at undergraduate calculus difficulty. Recent efforts address dataset contamination through rigorous decontamination in DeepMath-103K~\cite{he2025deepmath103klargescalechallengingdecontaminated} and functional variations in Putnam-MATH~\cite{tsoukalas2024putnambench}. While existing methods employ circular evaluation and multi-solution consistency checks, no prior work has simultaneously utilized both numerical and symbolic solutions for integral verification, limiting evaluation comprehensiveness.

\noindent\textbf{LLM-based evaluation.} Recent studies~\cite{chen2024mllm, deng-etal-2024-enabling, 10.5555/3666122.3668142, Bavaresco2024JUDGE_BENCH} employ LLMs as judges through designed prompts for scoring and comparisons, offering fast, scalable evaluation with GPT-4 achieving over 80\% agreement with human preferences~\cite{10.5555/3666122.3668142}. However, LLMs may inherit training biases and struggle with complex domain understanding~\cite{deng-etal-2024-enabling}. U-MATH~\cite{chernyshev2025umathuniversitylevelbenchmarkevaluating} proposes meta-evaluation to reduce bias, while our work combines LLM evaluation with external verification tools to maintain scalability while mitigating limitations.

\noindent\textbf{Mathematical reasoning models.} Techniques like chain-of-thought prompting~\cite{10.5555/3600270.3602070} and scaling instruction fine-tuning~\cite{10.5555/3722577.3722647} significantly improve LLM reasoning performance, while specialized models like Minerva~\cite{10.5555/3600270.3600548} achieve state-of-the-art results through training on scientific content. Mathematical reasoning requires combining natural language understanding, formula recall, and step-by-step calculation, making domain-specific benchmarks like \bench~essential for training specialized models and establishing performance standards that guide architectural improvements.
% Conclusion section for INTEGRALBENCH paper
\section{Discussion and Conclusion}

We introduced \bench, a specialized benchmark comprising 317 graduate-level definite integral problems with both symbolic and numerical ground truth solutions and manual difficulty annotations. 
Our evaluation of nine state-of-the-art LLMs reveals that while larger models generally perform better, architecture and training methodology are equally critical—the 32B QwQ model outperforms larger models like GPT-4.1 and Claude 3.7. 
We observe a strong negative correlation between problem difficulty and accuracy across all models, with performance declining sharply on challenging problems, validating our difficulty stratification. 
Our analysis of inference-time scaling shows models exhibit rapid early gains followed by plateauing, with distinct efficiency patterns across different architectures. 
Our failure mode analysis reveals critical weaknesses including output truncation, circular reasoning patterns, and format violations despite structured prompts. 
\bench~ provides a rigorous framework for evaluating advanced mathematical reasoning and serves as a valuable tool for guiding future architectural improvements in mathematical LLMs.

\bench~, while providing a robust foundation for evaluating LLM performance on definite integrals, has scope for refinement and expansion.
First, while the 317-problem dataset provides a foundational framework, its size remains relatively small for comprehensive evaluation.
To address this, future work will focus on developing more automated expansion methods—such as advanced OCR, LLM-assisted problem generation, and automated verification—to scale the dataset efficiently while maintaining mathematical rigor and quality.
Second, beyond evaluation, the dataset can serve as a training resource to fine-tune LLMs or specialized agents for enhanced mathematical reasoning.
Leveraging its structured problems and verified solutions could optimize models for integral computation specifically.
Third, integrating external tools—such as Lean (formal proofs), SymPy (symbolic computation), or Maple—with LLMs could augment their capabilities.
This could involve tool-augmented pipelines or even an "Integral Agent" to address current limitations in complex manipulation and precision.
Finally, while PASS@3 offers insights, expanding to PASS@16 would better capture LLMs’ peak performance, accounting for inference randomness and providing a more robust measure of their potential.
% Imapact statement for INTEGRALBENCH paper
\section*{Impact Statement}
% This paper presents work whose goal is to advance the field of Machine Learning. 
% There are many potential societal consequences of our work, none of which we feel must be specifically highlighted here.

The evaluation of large language models (LLMs) in advanced mathematical reasoning, particularly in definite integral computation, has long lacked a specialized and rigorous benchmark.
Existing mathematical benchmarks either contain insufficient integral problems, lack targeted evaluation metrics, or fail to implement effective difficulty stratification.
In this work, we introduce \bench, a focused benchmark comprising 317 graduate-level definite integral problems with both symbolic and numerical ground truth solutions, along with manual difficulty annotations.
By providing a dedicated framework for assessing integral computation capabilities, \bench~ fills the gap in current evaluation systems.
This work not only establishes baseline metrics for LLM performance in definite integral problems but also offers guidance for future research on improving mathematical reasoning in LLMs, enabling scholars and engineers to develop more robust models with enhanced symbolic and numerical reasoning abilities, and ultimately advancing the field of automated mathematical reasoning.
% Acknowledgement statement for INTEGRALBENCH paper
\section*{Acknowledgements}

This project is an independent work and has not received any external funding.
We would like to express our sincere gratitude to the anonymous reviewers for their insightful comments and constructive suggestions, which have significantly helped improve the quality of this work.
Additionally, we appreciate the efforts of the organizers of the second AI for MATH Workshop at the $\mathit{42}^{nd}$ International Conference on Machine Learning for providing a platform for communication and discussion.

\nocite{langley00}

% \bibliography{refs}

\bibliographystyle{icml2025}

%%%%%%%%%%%%%%%%%%%%%%%%%%%%%%%%%%%%%%%%%%%%%%%%%%%%%%%%%%%%%%%%%%%%%%%%%%%%%%%
%%%%%%%%%%%%%%%%%%%%%%%%%%%%%%%%%%%%%%%%%%%%%%%%%%%%%%%%%%%%%%%%%%%%%%%%%%%%%%%
% APPENDIX
%%%%%%%%%%%%%%%%%%%%%%%%%%%%%%%%%%%%%%%%%%%%%%%%%%%%%%%%%%%%%%%%%%%%%%%%%%%%%%%
%%%%%%%%%%%%%%%%%%%%%%%%%%%%%%%%%%%%%%%%%%%%%%%%%%%%%%%%%%%%%%%%%%%%%%%%%%%%%%%
\newpage
\appendix
\onecolumn
\section{Data Format}
The processed data is stored in the same JSON format for convenient use in the subsequent steps. An example is shown below:

\begin{lstlisting}[breaklines=true]
{
    "problem": "\\int\\limits_{0}^{1}\\frac{(1-x^2)^{-1/2}}{x(1-0.25x^2)^{1/2}}\\arcsin\\left(0.5x\\right)dx",
    "unsimplified_answer": "\\frac\\pi4\\ln\\frac{1+0.5}{1-0.5}",
    "simplified_answer": "\\frac\\pi4\\ln3",
    "numerical_answer": 0.8628480738,
    "difficulty": 4.0,
    "source": "Handbook_of_Special_Functions_in_page_156",
    "problem_number": "103"
}
\end{lstlisting}

\section{Prompt Template}

The following are some of the prompt templates used in our experiment:

\begin{lstlisting}[caption=System prompt.]
system_prompt: |
  You are a mathematical assistant skilled in generating Python code using mpmath to compute numerical results from analytical expressions. Your task is to:
  1. Parse the given analytical answer (NOT the integral) and translate it into mpmath code.
  2. Generate Python code that directly evaluates this analytical expression to exactly 10 decimal places.
  3. The code must ONLY print a single floating-point number (rounded to 10 decimal places). No other text or multiple outputs are allowed.
  4. IMPORTANT: Use mp.nstr(result, n=10) for output formatting, NOT Python string formatting like f'{result:.10f}' which doesn't work correctly with mpmath's mpf objects.
  5. Avoid using mpmath.quad or any numerical integration - instead, directly compute the value from the analytical formula.
  6. For expressions with parameters, select appropriate parameter values within the given constraints.
  7. For expressions with parameters, compute the numerical result by substituting the chosen parameter values into the given analytical answer.
  8. Use mpmath.dps = 15 for internal precision.
  9. IMPORTANT: Break complex calculations into multiple steps with intermediate variables, instead of doing everything in one line. This helps avoid syntax errors and makes the code more readable and debuggable.
  10. IMPORTANT: Carefully check your selected parameter values to ensure they do not cause division by zero, undefined logarithms, or other mathematical errors. Verify all denominators and logarithm arguments will be non-zero and positive respectively.
  11. EXTREMELY IMPORTANT: Always use the EXACT correct mpmath function names with the 'mp.' prefix. Common mistakes to avoid:
     - Use mp.asin(x), NOT mp.arcsin(x) - arcsin is not a valid mpmath function
     - Use mp.acos(x), NOT mp.arccos(x) - arccos is not a valid mpmath function 
     - Use mp.atan(x), NOT mp.arctan(x) - arctan is not a valid mpmath function
     - Use mp.log(x), NOT mp.ln(x) - ln is not a valid mpmath function
     - Use mp.ellipe(x), NOT mp.ellipd(x) - ellipd is not a valid mpmath function
     - Use mp.struveh(v, x) for the Hankel Struve function \mathbf{H}_v(z)
     - Use mp.struvel(v, x) for the modified Struve function \mathbf{L}_v(z)
     - Use mp.sin(x), NOT math.sin(x) or sin(x)
     - Use mp.exp(x), NOT math.exp(x) or exp(x)
     - Use mp.sqrt(x), NOT math.sqrt(x)
     - Use mp.pi, NOT math.pi or PI
     - Use mp.e, NOT math.e or e
  12. Return a JSON object with:
     - "code": The Python code (plain text, no markdown fences) that evaluates the analytical expression.
     - "parameters": A dictionary of chosen parameter values (for parameter cases only; null for non-parameter cases).
     - "substituted_answer": The analytical answer with parameters substituted, in LaTeX format (before simplification).
     - "simplified_answer": ONLY the final simplified expression in LaTeX format without any intermediate steps or equals signs. Examples:
        - Good: "\\frac{4}{3}" (only the final result)
        - Bad: "\\frac{8}{6} = \\frac{4}{3}" (contains equals sign and intermediate step)
        - Good: "2\\pi" (direct simplified result)
        - Bad: "\\frac{2\\pi}{\\sqrt{1}} = 2\\pi" (contains intermediate step)
     - "substituted_problem": The integral expression with parameters substituted (for parameter cases only; null for non-parameter cases).

user_prompt_with_param: |
  Given the integral: ${problem}
  Parameter constraints: ${param}
  Analytical answer: ${answer}
  Generate a JSON object containing:
  - "code": Python code using mpmath to DIRECTLY evaluate the analytical answer to 10 decimal places. The code must only print a single number.
  - "parameters": Dictionary of chosen parameter values satisfying the constraints.
  - "substituted_answer": Analytical answer with parameters substituted, in LaTeX format (before simplification).
  - "simplified_answer": ONLY the final simplified expression without any intermediate steps or equals signs. Do not include expressions like "\\frac{8}{6} = \\frac{4}{3}" - just provide the final result "\\frac{4}{3}".
  - "substituted_problem": Integral expression with parameters substituted, in LaTeX format.

  Code quality requirements:
  1. Do NOT use numerical integration. Instead, translate the analytical answer into mpmath code and evaluate it directly.
  2. Break complex expressions into smaller, manageable parts using intermediate variables.
  3. Do not compute everything in a single line - use multiple steps to avoid errors.
  4. Include comments explaining each significant calculation step.
  5. Verify all denominators and logarithm arguments to ensure no mathematical errors.
  6. Use the EXACT correct mpmath functions with the 'mp.' prefix:
     - Use mp.asin(x), NOT mp.arcsin(x)
     - Use mp.acos(x), NOT mp.arccos(x) 
     - Use mp.atan(x), NOT mp.arctan(x)
     - Use mp.log(x), NOT mp.ln(x)
     - Use mp.ellipe(x), NOT mp.ellipd(x)
     - Use mp.struveh(v, x) for the Hankel Struve function \mathbf{H}_v(z)
     - Use mp.struvel(v, x) for the modified Struve function \mathbf{L}_v(z)
  7. For printing the result, use ONLY: print(mp.nstr(result, n=10))

user_prompt_without_param: |
  Given the integral: ${problem}
  Analytical answer: ${answer}
  Generate a JSON object containing:
  - "code": Python code using mpmath to DIRECTLY evaluate the analytical answer to 10 decimal places (without using numerical integration). The code must only print a single number.
  - "parameters": null
  - "substituted_answer": null
  - "simplified_answer": null
  - "substituted_problem": null

  Code quality requirements:
  1. Do NOT use mpmath.quad or any numerical integration. Instead, translate the analytical answer into mpmath code.
  2. Break complex expressions into smaller, manageable parts using intermediate variables.
  3. Do not compute everything in a single line - use multiple steps to avoid errors.
  4. Include comments explaining each significant calculation step.
  5. Use the EXACT correct mpmath functions with the 'mp.' prefix:
     - Use mp.asin(x), NOT mp.arcsin(x)
     - Use mp.acos(x), NOT mp.arccos(x) 
     - Use mp.atan(x), NOT mp.arctan(x)
     - Use mp.log(x), NOT mp.ln(x)
     - Use mp.ellipe(x), NOT mp.ellipd(x)
     - Use mp.struveh(v, x) for the Hankel Struve function \mathbf{H}_v(z)
     - Use mp.struvel(v, x) for the modified Struve function \mathbf{L}_v(z)
  6. For printing the result, use ONLY: print(mp.nstr(result, n=10))
\end{lstlisting}

\begin{lstlisting}[caption=Integral prompt.]
integral_prompt: |
  Please calculate the definite integral ${problem}.
  Follow these steps:
  1. Compute the integral analytically to obtain the exact answer.
  2. Provide all necessary steps to reach the solution.
  3. Include a numerical approximation of the result, rounded to 10 decimal places.
  4. Output the final answer in the following JSON format on the last line, with no additional text:
  {"answer": "exact_answer_in_latex", "numerical_answer": "numerical_approximation_of_the_result"}
\end{lstlisting}

\section{Full Failure Cases}

\begin{lstlisting}[caption=Output truncation.]
To evaluate the definite integral

$$
\int_0^{2} x^{-3/4} (2 - x)^{-1/4} I_1\left(\sqrt[4]{x(2 - x)}\right) \, dx,
$$

we begin by analyzing the structure of the integrand. It involves powers of $ x $ and $ 2 - x $, and a modified Bessel function $ I_1 $ of the fourth root of the product $ x(2 - x) $. This suggests that the integral might be expressible in terms of special functions or known identities involving Bessel functions.

---

### Step 1: Substitution and Series Expansion

We start by using the substitution $ x = 2t $, which transforms the integral into:

$$
\int_0^1 t^{-3/4} (1 - t)^{-1/4} I_1\left(\sqrt{2} \cdot \sqrt[4]{t(1 - t)}\right) dt.
$$

We then expand the modified Bessel function $ I_1(z) $ using its series representation:

$$
I_1(z) = \sum_{k=0}^\infty \frac{(z/2)^{2k+1}}{k! (k+1)!}.
$$

Substituting $ z = \sqrt{2} \cdot \sqrt[4]{t(1 - t)} $, we obtain:

$$
I_1\left(\sqrt{2} \cdot \sqrt[4]{t(1 - t)}\right) = \sum_{k=0}^\infty \frac{(1)^{2k+1}}{k! (k+1)!} \cdot \left(\sqrt{t(1 - t)}\right)^{2k+1}.
$$

Substituting this into the integral and interchanging the sum and integral (justified by uniform convergence), we get:

$$
\int_0^1 t^{-3/4} (1 - t)^{-1/4} \sum_{k=0}^\infty \frac{\left(\sqrt{t(1 - t)}\right)^{2k+1}}{k! (k+1)!} dt
= \sum_{k=0}^\infty \frac{1}{k! (k+1)!} \int_0^1 t^{-3/4 + k + 1/2} (1 - t)^{-1/4 + k + 1/2} dt.
$$

Simplifying the exponents:

$$
t^{-3/4 + k + 1/2} = t^{k - 1/4}, \quad (1 - t)^{-1/4 + k + 1/2} = (1 - t)^{k + 1/4}.
$$

Thus, the integral becomes:

$$
\sum_{k=0}^\infty \frac{1}{k! (k+1)!} \int_0^1 t^{k - 1/4} (1 - t)^{k + 1/4} dt.
$$

This is a standard Beta function integral:

$$
\int_0^1 t^{a-1} (1 - t)^{b-1} dt = B(a, b) = \frac{\Gamma(a)\Gamma(b)}{\Gamma(a + b)}.
$$

So the integral becomes:

$$
\sum_{k=0}^\infty \frac{\Gamma(k + 3/4)\Gamma(k + 5/4)}{k! (k+1)! \Gamma(2k + 2)}.
$$

This series converges rapidly and can be evaluated numerically to high precision.

---

### Step 2: Numerical Approximation

Computing the first few terms of the series gives:

- $ k = 0 $: $ \frac{\Gamma(3/4)\Gamma(5/4)}{0! \cdot 1! \cdot \Gamma(2)} = \frac{\Gamma(3/4)\Gamma(5/4)}{1} \approx 2 $
- $ k = 1 $: $ \frac{\Gamma(7/4)\Gamma(9/4)}{1! \cdot 2! \cdot \Gamma(4)} \approx 1/6 $
- $ k = 2 $: $ \frac{\Gamma(11/4)\Gamma(13/4)}{2! \cdot 3! \cdot \Gamma(6)} \approx 1/180 $

Higher terms decay rapidly. Summing the first few terms and multiplying by
\end{lstlisting}

\begin{lstlisting}[caption=Circular reasoning patterns.]
To solve the definite integral \(\int\limits_0^2 x^{1/2} I_0\left(1 \cdot \sqrt[4]{x(2-x)}\right) dx\), we need to carefully analyze the integrand and use appropriate techniques. Here, \(I_0\) is the modified Bessel function of the first kind of order zero.

### Step 1: Analytical Computation

The integrand is \(x^{1/2} I_0\left(\sqrt[4]{x(2-x)}\right)\). This integral does not have a straightforward antiderivative, and it involves a special function. Therefore, we will use a series expansion for \(I_0\) and then integrate term by term.

The modified Bessel function of the first kind of order zero, \(I_0(z)\), has the series expansion:
\[ I_0(z) = \sum_{k=0}^{\infty} \frac{1}{k! \, k!} \left(\frac{z}{2}\right)^{2k}. \]

Substituting \(z = \sqrt[4]{x(2-x)}\), we get:
\[ I_0\left(\sqrt[4]{x(2-x)}\right) = \sum_{k=0}^{\infty} \frac{1}{k! \, k!} \left(\frac{\sqrt[4]{x(2-x)}}{2}\right)^{2k}. \]

Thus, the integrand becomes:
\[ x^{1/2} \sum_{k=0}^{\infty} \frac{1}{k! \, k!} \left(\frac{\sqrt[4]{x(2-x)}}{2}\right)^{2k}. \]

We can interchange the summation and integration (justified by uniform convergence):
\[ \int_0^2 x^{1/2} I_0\left(\sqrt[4]{x(2-x)}\right) dx = \sum_{k=0}^{\infty} \frac{1}{k! \, k!} \left(\frac{1}{2}\right)^{2k} \int_0^2 x^{1/2} \left(x(2-x)\right)^{k/2} dx. \]

Simplifying the integrand inside the sum:
\[ x^{1/2} \left(x(2-x)\right)^{k/2} = x^{1/2} x^{k/2} (2-x)^{k/2} = x^{(k+1)/2} (2-x)^{k/2}. \]

Thus, the integral becomes:
\[ \int_0^2 x^{(k+1)/2} (2-x)^{k/2} dx. \]

This integral is a form of the Beta function, \(B(a, b)\), which is defined as:
\[ B(a, b) = \int_0^1 t^{a-1} (1-t)^{b-1} dt. \]

To match this form, we make a substitution \(t = \frac{x}{2}\), so \(x = 2t\) and \(dx = 2 dt\). The limits change from \(x = 0\) to \(x = 2\) to \(t = 0\) to \(t = 1\). The integral becomes:
\[ \int_0^1 (2t)^{(k+1)/2} (2-2t)^{k/2} 2 dt = 2^{(k+1)/2 + k/2 + 1} \int_0^1 t^{(k+1)/2} (1-t)^{k/2} dt. \]

Simplifying the powers of 2:
\[ 2^{(k+1)/2 + k/2 + 1} = 2^{(k+1)/2 + k/2 + 1} = 2^{(k+1)/2 + k/2 + 1} = 2^{(k+1)/2 + k/2 + 1} = 2^{(k+1)/2 + k/2 + 1} = 2^{(k+1)/2 + k/2 + 1} = 2^{(k+1)/2 + k/2 + 1} = 2^{(k+1)/2 + k/2 + 1} = 2^{(k+1)/2 + k/2 + 1} = 2^{(k+1)/2 + k/2 + 1} = 2^{(k+1)/2 + k/2 + 1} = 2^{(k+1)/2 + k/2 + 1} = 2^{(k+1)/2 + k/2 + 1} = 2^{(k+1)/2 + k/2 + 1}
....(repeating ' = 2^{(k+1)/2 + k/2 + 1}')
\end{lstlisting}

\begin{lstlisting}[caption=Format violations.]
To solve the definite integral \(\int\limits_{0}^{1} x \mathbf{K}\left(\sqrt{x}\right) \mathbf{K}\left(\sqrt{1-x}\right) dx\), we need to use properties of the complete elliptic integral of the first kind, \(\mathbf{K}(k)\).

### Step 1: Analytical Computation

The complete elliptic integral of the first kind is defined as:
\[
\mathbf{K}(k) = \int_0^{\frac{\pi}{2}} \frac{1}{\sqrt{1 - k^2 \sin^2 \theta}} \, d\theta
\]

We need to evaluate the integral:
\[
I = \int_0^1 x \mathbf{K}\left(\sqrt{x}\right) \mathbf{K}\left(\sqrt{1-x}\right) dx
\]

Using the known result for the product of two complete elliptic integrals of the first kind, we have:
\[
\mathbf{K}(k) \mathbf{K}(\sqrt{1-k^2}) = \frac{\pi^2}{4}
\]

However, this identity does not directly help us with the integral involving \(x\). Instead, we use a known result from the literature on elliptic integrals:
\[
\int_0^1 x \mathbf{K}\left(\sqrt{x}\right) \mathbf{K}\left(\sqrt{1-x}\right) dx = \frac{\pi^4}{16}
\]

### Step 2: Numerical Approximation

To provide a numerical approximation of \(\frac{\pi^4}{16}\), we first compute \(\pi^4\):
\[
\pi \approx 3.141592653589793
\]
\[
\pi^4 \approx 97.40909103400244
\]
\[
\frac{\pi^4}{16} \approx \frac{97.40909103400244}{16} \approx 6.088068189625152
\]

Rounded to 10 decimal places, the numerical approximation is:
\[
6.0880681896
\]

### Step 3: Final Answer in JSON Format

\[
\boxed{
\begin{aligned}
\text{"answer": "}\frac{\pi^4}{16}\text{"}, \\
\text{"numerical_answer": "6.0880681896"}
\end{aligned}
}
\]
\end{lstlisting}

\begin{lstlisting}[caption=Refusal to provide symbolic answers.]
To solve the definite integral \(\int\limits_{0}^{2} x^{-1/4}(2-x)^{-3/4} I_{0.5}\left(1 \cdot \sqrt[4]{x(2-x)}\right) dx\), we need to carefully analyze the integrand and use appropriate techniques.

### Step 1: Analytical Solution

The integrand involves the modified Bessel function of the first kind, \(I_{0.5}\). The modified Bessel function \(I_{\nu}(z)\) for \(\nu = 0.5\) can be expressed in terms of elementary functions:
\[ I_{0.5}(z) = \sqrt{\frac{2}{\pi z}} \sinh(z). \]

Given \(z = \sqrt[4]{x(2-x)}\), we have:
\[ I_{0.5}\left(\sqrt[4]{x(2-x)}\right) = \sqrt{\frac{2}{\pi \sqrt[4]{x(2-x)}}} \sinh\left(\sqrt[4]{x(2-x)}\right). \]

Thus, the integrand becomes:
\[ x^{-1/4}(2-x)^{-3/4} \sqrt{\frac{2}{\pi \sqrt[4]{x(2-x)}}} \sinh\left(\sqrt[4]{x(2-x)}\right). \]

Simplifying the expression inside the integral:
\[ x^{-1/4}(2-x)^{-3/4} \sqrt{\frac{2}{\pi \sqrt[4]{x(2-x)}}} \sinh\left(\sqrt[4]{x(2-x)}\right) = \frac{\sqrt{2}}{\sqrt{\pi}} \cdot \frac{\sinh\left(\sqrt[4]{x(2-x)}\right)}{x^{1/8}(2-x)^{3/8}}. \]

This integral is quite complex and does not have a straightforward analytical solution. Therefore, we will proceed to a numerical approximation.

### Step 2: Numerical Approximation

To numerically approximate the integral, we can use numerical integration techniques such as the trapezoidal rule, Simpson's rule, or more advanced methods like Gaussian quadrature. For simplicity, we will use a numerical integration tool available in most computational software.

Using a numerical integration tool, we find:
\[ \int\limits_{0}^{2} x^{-1/4}(2-x)^{-3/4} I_{0.5}\left(1 \cdot \sqrt[4]{x(2-x)}\right) dx \approx 1.7724538509. \]

### Step 3: Final Answer in JSON Format

Since the integral does not have a simple closed-form solution, we provide the numerical approximation as the final answer.

```json
{"answer": "\\text{No simple closed-form solution}", "numerical_answer": "1.7724538509"}
```
\end{lstlisting}

\begin{lstlisting}[caption=Symbolic-numerical inconsistency.]
Let's compute the integral:
\[
I = \int_0^{0.5} x(0.5^2 - x^2)^{-1/2} \arcsin(0.5x) \, dx
\]

**Step 1: Substitution**

Let'ss use the substitution:
\[
x = 0.5 \sin\theta \implies dx = 0.5 \cos\theta\, d\theta
\]
As \(x\) goes from \(0\) to \(0.5\):
- When \(x = 0: \sin\theta = 0 \implies \theta = 0\)
- When \(x = 0.5: \sin\theta = 1 \implies \theta = \frac{\pi}{2}\)

Calculate \((0.5^2 - x^2)^{-1/2}\):
\[
0.5^2 - x^2 = 0.25 - 0.25\sin^2\theta = 0.25(1 - \sin^2\theta) = 0.25\cos^2\theta
\]
\[
(0.5^2 - x^2)^{-1/2} = (0.25 \cos^2\theta)^{-1/2} = 2 \sec\theta
\]

Also,
\[
x = 0.5 \sin\theta
\]
\[
0.5x = 0.25\sin\theta
\]
\[
\arcsin(0.5x) = \arcsin(0.25\sin\theta)
\]

Substitute everything:
\[
I = \int_{0}^{\frac{\pi}{2}} [0.5 \sin\theta][2\sec\theta]\arcsin(0.25\sin\theta)[0.5\cos\theta] d\theta
\]
\[
= \int_{0}^{\frac{\pi}{2}} 0.5 \cdot 2 \cdot 0.5 \sin\theta \sec\theta \arcsin(0.25\sin\theta)\cos\theta\, d\theta
\]
\[
= 0.5 \int_{0}^{\frac{\pi}{2}} \sin\theta \sec\theta \arcsin(0.25\sin\theta)\cos\theta \, d\theta
\]
Note that \(\sec\theta \cos\theta = 1\). Therefore,
\[
I = 0.5 \int_0^{\frac{\pi}{2}} \sin\theta \arcsin(0.25\sin\theta) d\theta
\]

**Step 2: Using Integral by Parts**

Let \(u = \arcsin(0.25\sin\theta)\), \(dv = \sin\theta d\theta\), so \(du = \frac{0.25\cos\theta}{\sqrt{1-(0.25\sin\theta)^2}} d\theta\), and \(v = -\cos\theta\).

Using integration by parts:
\[
\int u\, dv = uv - \int v\, du
\]
So,
\[
I = 0.5 \left\{ -\cos\theta \arcsin(0.25\sin\theta) \Big|_0^{\frac{\pi}{2}} + \int_0^{\frac{\pi}{2}} \cos\theta \cdot \frac{0.25\cos\theta}{\sqrt{1-0.0625\sin^2\theta}} d\theta \right\}
\]
\[
= 0.5 \left\{ -\cos\theta \arcsin(0.25\sin\theta) \Big|_0^{\frac{\pi}{2}} + 0.25\int_0^{\frac{\pi}{2}} \frac{\cos^2\theta}{\sqrt{1-0.0625\sin^2\theta}} d\theta \right\}
\]

Compute the boundary term:
- At \(\theta = \frac{\pi}{2}\): \(\cos\theta = 0\), so term is 0.
- At \(\theta = 0\): \(\cos\theta = 1, \sin\theta = 0 \implies \arcsin(0)=0\), so term is 0.

Therefore, the boundary term is zero.

So,
\[
I = 0.5 \times 0.25 \int_0^{\frac{\pi}{2}} \frac{\cos^2\theta}{\sqrt{1-0.0625\sin^2\theta}} d\theta
= 0.125 \int_0^{\frac{\pi}{2}} \frac{\cos^2\theta}{\sqrt{1-0.0625\sin^2\theta}} d\theta
\]

**Step 3: Simplifying the Integral**

Recall:
\[
\cos^2\theta = 1 - \sin^2\theta
\]
So,
\[
I = 0.125 \int_0^{\frac{\pi}{2}} \frac{1 - \sin^2\theta}{\sqrt{1 - 0.0625\sin^2\theta}} d\theta
= 0.125 \left(
\int_0^{\frac{\pi}{2}} \frac{d\theta}{\sqrt{1 - 0.0625\sin^2\theta}}
- \int_0^{\frac{\pi}{2}} \frac{\sin^2\theta\, d\theta}{\sqrt{1 - 0.0625\sin^2\theta}}
\right)
\]

**Step 4: Express in terms of Elliptic Integrals**

We use the complete elliptic integrals of the first and second kind, defined as:
\[
K(k) = \int_0^{\frac{\pi}{2}} \frac{d\theta}{\sqrt{1 - k^2\sin^2\theta}}
\]
\[
E(k) = \int_0^{\frac{\pi}{2}} \sqrt{1 - k^2\sin^2\theta}\, d\theta
\]
And:
\[
\int_0^{\frac{\pi}{2}} \frac{\sin^2\theta}{\sqrt{1 - k^2\sin^2\theta}} d\theta = \frac{1}{k^2}\left[ K(k) - E(k) \right]
\]
Let \(k^2 = 0.0625\), \(k = 0.25\).

So,
\[
I = 0.125\left[ K(0.25) - \frac{1}{0.0625}(K(0.25) - E(0.25)) \right]
= 0.125\left[ K(0.25) - 16(K(0.25)-E(0.25)) \right]
= 0.125\left[ K(0.25) - 16K(0.25) + 16E(0.25) \right]
= 0.125\left[ -15K(0.25) + 16E(0.25) \right]
\]
\[
I = 0.125 \left( 16E(0.25) - 15K(0.25) \right)
\]

**Step 5: Final Analytical Answer**

Thus,
\[
I = 0.125 \left[ 16E(0.25) - 15K(0.25) \right]
\]

**Step 6: Numerical Approximation**

Let's calculate the values:
- \(K(0.25) \approx 1.5962422221\)
- \(E(0.25) \approx 1.5307576369\)

So,
\[
16E(0.25) - 15K(0.25) \approx 16 \times 1.5307576369 - 15 \times 1.5962422221 = 24.49212219 - 23.9436333315 = 0.5484888585
\]
\[
I \approx 0.125 \times 0.5484888585 = 0.0685611073
\]

**Step 7: Final JSON**

```json
{"answer": "0.125\\left[16\\,E(0.25)-15\\,K(0.25)\\right]", "numerical_answer": "0.0685611073"}
```
\end{lstlisting}

% \onecolumn

% You can have as much text here as you want. The main body must be at most $8$ pages long.
% For the final version, one more page can be added.
% If you want, you can use an appendix like this one.  

% The $\mathtt{\backslash onecolumn}$ command above can be kept in place if you prefer a one-column appendix, or can be removed if you prefer a two-column appendix.  Apart from this possible change, the style (font size, spacing, margins, page numbering, etc.) should be kept the same as the main body.
% %%%%%%%%%%%%%%%%%%%%%%%%%%%%%%%%%%%%%%%%%%%%%%%%%%%%%%%%%%%%%%%%%%%%%%%%%%%%%%%
% %%%%%%%%%%%%%%%%%%%%%%%%%%%%%%%%%%%%%%%%%%%%%%%%%%%%%%%%%%%%%%%%%%%%%%%%%%%%%%%

\end{document}